\documentclass{article}

\usepackage[english]{babel}

\usepackage[letterpaper,top=2cm,bottom=2cm,left=3cm,right=3cm,marginparwidth=1.75cm]{geometry}
\usepackage{authblk}

\usepackage{amsmath}
\usepackage{graphicx}
\usepackage{subcaption}
\usepackage[toc,page]{appendix}
\usepackage{float}
\usepackage[colorlinks=true, allcolors=blue]{hyperref}

\title{The rUNSWift SPL Field Segmentation Dataset}
\author{\textbf{Wentao Lu}}
\affil{School of Computer Science and Engineering\\
The University of New South Wales\\
wentao.lu@unsw.edu.au}

\begin{document}
\maketitle

\begin{abstract}
In RoboCup SPL, soccer field segmentation has been widely recognised as one of the most critical robot vision problems. Key challenges include dynamic light condition, different calibration status for individual robot, various camera prospective and more. In this paper, we propose a dataset that contains 20 videos recorded with Nao V5/V6 humanroid robots\footnote{https://www.softbankrobotics.com/emea/en/nao} by team rUNSWift under different circumstances. Each of the videos contains several consecutive high resolution frames and the corresponding labels for field. We propose this dataset to provide training data for the league to overcome field segmentation problem.
The dataset will be available online for download. Details of annotation and example of usage will be explained in later sections.
\end{abstract}

\section{Introduction}
Since the beginning of RoboCup SPL competition, field segmentation/recognition remains a key problem of robot vision system. In early years, field line detection with RANSAC\cite{10.1145/358669.358692} based field recognition is widely used. This method can roughly find the field area as per the rule, field lines are 50 centimetres away from the field boundary. This method works but one critical disadvantage is in some cases, the robot is not able to see the field lines. In late years, many teams have proposed colour detection based field recognition\cite{tjark} which appears to be more robust and flexible to use. To detect a certain colour, the method must overcome several challenges. For example, since 2016, our league has introduced outdoor competitions where the light condition varies and could affect the colour detection\cite{SPL016}. Recently, thanks to the introduction of Nao V6 robot, the on-board computation power enriches which allows us to run more complicatied learning algorithms to recognise ball, robots, and also the soccer field. Learning algorithms require large amount of data. While many teams have kindly shared the ball and robot image dataset, datasets for field segmentation remains few. To make contribution to this particular problem, we propose this dataset and corresponding labels.\\
The complete dataset can be downloaded from \url{https://drive.google.com/file/d/17EDkVGvqW46Cn8a7\_9uGuXDlqvweOvE4/view?usp=sharing}.

\section{Dataset Specifications}
In this section, we will explain how the data look like and how we collect the data. In Section 2.1, we will show the format and other specifications of the dataset. Data collection details will be discussed in Section 2.2. 

\subsection{Dataset format}
Table \ref{Dataset specs} shows the data format and resolution for each video in the dataset. Our dataset contains 7 short videos and 13 longer ones. Each video is decomposed into consecutive image frames where the image format is listed in Table \ref{Dataset specs} as well. Due to the various situation when we collect the data, only short videos have bottom frames attached while the longer ones we only keep top frames as field segmentation is required mainly on top camera. Resolution for top camera remains 1280 * 960 while the corresponding label is downsampled by 5 times to 256 * 192. Resolution for bottom images of shorter videos is halfed to 640 * 480. Raw image format from the robot is YUV and for most videos in our dataset, the image frames have been converted to BMP format.

\begin{table}[!h]
\begin{tabular}{|l|l|l|l|l|l|}
\hline
Index & Resolution(Top) & Label(Top) & Resolution(Bottom) & Image Format & \#Frames \\ \hline
1     & 1280 * 960      & 256 * 192  & 640 * 480          & YUV          & 19       \\ \hline
2     & 1280 * 960      & 256 * 192  & 640 * 480          & YUV          & 19       \\ \hline
3     & 1280 * 960      & 256 * 192  & 640 * 480          & YUV          & 19       \\ \hline
4     & 1280 * 960      & 256 * 192  & 640 * 480          & BMP          & 5        \\ \hline
5     & 1280 * 960      & 256 * 192  & 640 * 480          & BMP/YUV      & 6        \\ \hline
7     & 1280 * 960      & 256 * 192  & 640 * 480          & BMP/YUV      & 19       \\ \hline
8     & 1280 * 960      & 256 * 192  & 640 * 480          & BMP/YUV      & 7        \\ \hline
9     & 1280 * 960      & 256 * 192  & N/A                & BMP          & 259      \\ \hline
10    & 1280 * 960      & 256 * 192  & N/A                & BMP          & 199      \\ \hline
11    & 1280 * 960      & 256 * 192  & N/A                & BMP          & 121      \\ \hline
12    & 1280 * 960      & 256 * 192  & N/A                & BMP          & 271      \\ \hline
13    & 1280 * 960      & 256 * 192  & N/A                & BMP          & 113      \\ \hline
14    & 1280 * 960      & 256 * 192  & N/A                & BMP          & 68       \\ \hline
15    & 1280 * 960      & 256 * 192  & N/A                & BMP          & 119      \\ \hline
16    & 1280 * 960      & 256 * 192  & N/A                & BMP          & 298      \\ \hline
17    & 1280 * 960      & 256 * 192  & N/A                & BMP          & 251      \\ \hline
18    & 1280 * 960      & 256 * 192  & N/A                & BMP          & 210      \\ \hline
19    & 1280 * 960      & 256 * 192  & N/A                & BMP          & 181      \\ \hline
20    & 1280 * 960      & 256 * 192  & N/A                & BMP          & 213      \\ \hline
21    & 1280 * 960      & 256 * 192  & N/A                & BMP          & 188      \\ \hline
\end{tabular}
\caption{Dataset specs}
\label{Dataset specs}
\end{table}

\subsection{Data collection}
In this section, we will discuss how we collect those data in detail. We use Nao V5/V6 robots to run different behaviour in different conditions to collect data. When the robot running, the raw images, usually a video, will be recorded with other data. The combined data will be logged into a dump file which will later be downloaded to local machine. Team rUNSWift has two tools to record and decode the dump file, which are Offnao and Vatnao\cite{runswift}. More information about these tools can be founded in rUNSWift's team report. Vatnao loads saved dump file and decodes the video into consecutive frames which can be saved in selected image format.\\
Table \ref{Dataset information} shows location, light condition and behaviour setting for each video. 7 shorter videos were recorded in Germany when team rUNSWift participated in RoboCup German Open 2019. All those videos are recorded in indoor field with different light condition and robot behaviour. Longer videos were recorded in Sydney, Australia with 3 different locations which are UNSW robot laboratory, Internation Conference Centre's indoor field and outdoor field. Those data recorded in outdoor field also cover different time where we aim to include more natural and various light condition.\\
Table \ref{Dataset information} also includes body motion and recording type. Body motion explains whether the robot that is recording the dump file moves. Recording type demonstrates what condition is while we record the data. Single-testing means only one robot is operating on the field while recording data. Multi-testing means multiple robots are running where those robots can occupy parts of the field which is a harder case for field segmentation.

\begin{table}[!h]
\begin{tabular}{|l|l|l|l|l|l|}
\hline
\textbf {index} & \textbf{\#frame} & \textbf{location} & \textbf{light condition} & \textbf{body motion} & \textbf{recording type} \\ \hline
1 &19               & Indoor(Germany)   & Constant(low)            & Static               & Multi-testing           \\ \hline
2 &19               & Indoor(Germany)   & Constant(low)            & Static               & Multi-testing           \\ \hline
3 &19               & Indoor(Germany)   & Constant(low)            & Static               & Multi-testing           \\ \hline
4 &5                & Indoor(Germany)   & Constant(high)           & Static               & Single-testing          \\ \hline
5 &5                & Indoor(Germany)   & Constant(high)           & Static               & Single-testing          \\ \hline
7 &19               & Indoor(Germany)   & Constant(low)            & Dynamic              & Single-testing          \\ \hline
8 &7                & Indoor(Germany)   & Constant(low)            & Dynamic              & Game                    \\ \hline
9 &259              & Indoor(UNSW)      & Constant(medium)         & Dynamic              & Single-testing          \\ \hline
10 &199              & Outdoor(Sydney)   & Night                    & Static               & Single-testing          \\ \hline
11 &121              & Indoor(UNSW)      & Uneven(medium)           & Dynamic              & Single-testing          \\ \hline
12 &271              & Indoor(UNSW)      & Uneven(medium)           & Dynamic              & Multi-testing           \\ \hline
13 &113              & Indoor(UNSW)      & Uneven(medium)           & Dynamic              & Multi-testing           \\ \hline
14 &68               & Outdoor(Sydney)   & Night                    & Static               & Single-testing          \\ \hline
15 &119              & Outdoor(Sydney)   & Night                    & Static               & Single-testing          \\ \hline
16 &298              & Indoor(Sydney)    & Constant(low)            & Dynamic              & Multi-testing           \\ \hline
17 &251              & Outdoor(Sydney)   & Morning(uneven)          & Static               & Single-testing          \\ \hline
18 &210              & Indoor(UNSW)      & Extremely uneven         & Dynamic              & Single-testing          \\ \hline
19 &181              & Indoor(UNSW)      & Extremely uneven         & Dynamic              & Single-testing          \\ \hline
20 &213              & Indoor(UNSW)      & Extremely uneven         & Dynamic              & Multi-testing           \\ \hline
21 &188              & Indoor(UNSW)      & Extremely uneven         & Dynamic              & Multi-testing           \\ \hline
\end{tabular}
\caption{Dataset Information}
\label{Dataset information}
\end{table}

\section{Data Annotation}
In this section, we will discuss how we label the field and how to utilise the provided image label. In Section 3.2, a brief statistics for each video will be demonstrated. A simple tool will also be introduced in Section 3.3, which allows user to label their own image data.
\subsection{Labelling system}
\begin{figure}[!h]
	\centering
	\begin{subfigure}[b]{0.49\textwidth}
		\includegraphics[width=\textwidth]{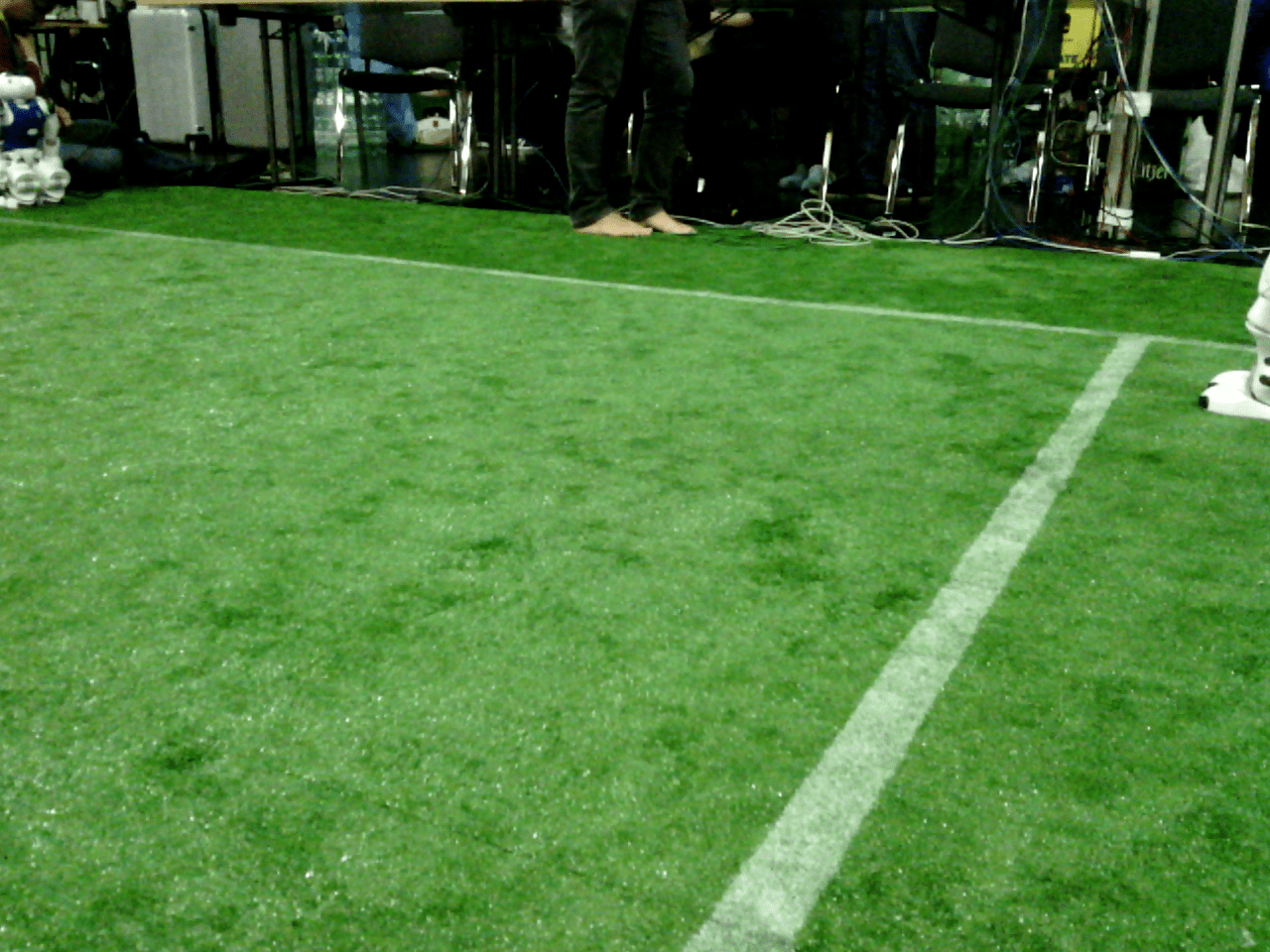}
		\caption{Raw Image}
		\label{fig:rawimage1}
	\end{subfigure}
	\begin{subfigure}[b]{0.49\textwidth}
		\includegraphics[width=\textwidth]{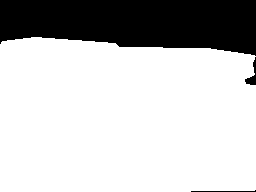}
		\caption{Label}
		\label{fig:label1}
	\end{subfigure}
	
	\caption{Data pair collected in Germany}
	\label{fig:labels1}
\end{figure}

\begin{figure}[!h]
	\centering
	\begin{subfigure}[b]{0.49\textwidth}
		\includegraphics[width=\textwidth]{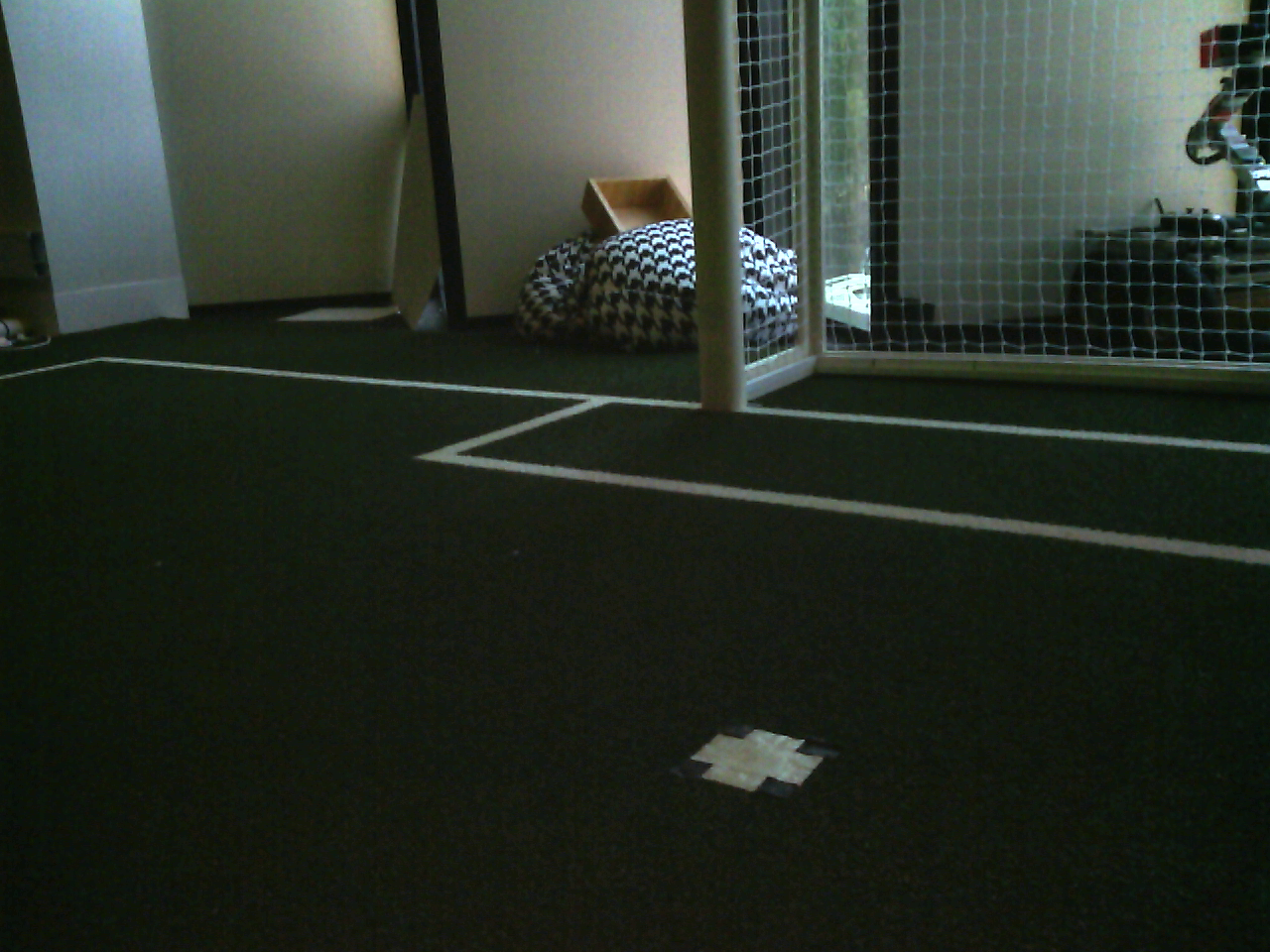}
		\caption{Raw Image}
		\label{fig:rawimage2}
	\end{subfigure}
	\begin{subfigure}[b]{0.49\textwidth}
		\includegraphics[width=\textwidth]{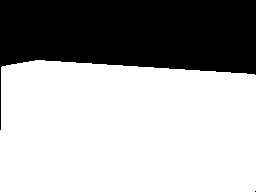}
		\caption{Label}
		\label{fig:label2}
	\end{subfigure}
	
	\caption{Data pair collected in UNSW}
	\label{fig:labels2}
\end{figure}

Figure \ref{fig:labels1} and \ref{fig:labels2} show two examples regarding how we label the field. The image label is a 256 * 192 monochrome picture which is downsampled by 5 times from full resolution(1280 * 960). Within the image label, the soccer field is marked as white pixels while all other stuff marked as black. Generally, lines within the soccer field are also marked with white pixels due to workload restrictions. 

\newpage
\subsection{Field Statistics}
In this section, we will run through a list of statistics to demonstrate how the data look like from a mathematical prospective.
\begin{figure}[!h]
	\centering
	\includegraphics[width=\textwidth]{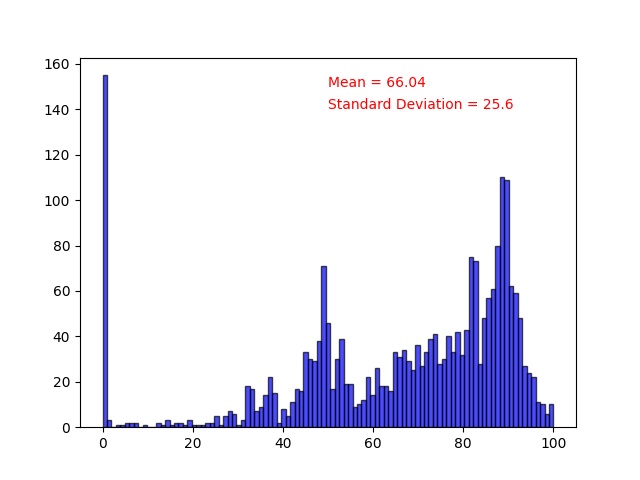}
	\caption{Statistics for soccer field coverage}
	\label{fig:stats1}
\end{figure}
Figure \ref{fig:stats1} shows a statistics for coverage of soccer field of the whole dataset. As the statistics shown, most images have a field coverage around 50\% and 90\% where the average value is 66\%. 155 images of total 2390 are field-less(no field covered) with pure black labels. Individual statistics for each video can be found in Appendix \ref{appendix:stats2}.

\subsection{Tools}
A labelling tool is provided along with the dataset. This tool is written in Python 3 with OpenCV2. When given the image folder path, the function will load the image and start a mouse click tracking procedure. By clicking left mouse button, user can draw key points on the image. If all key points marked, user can click middle mouse button to show the connected component created by those key points which is the labelled field. If user satisfies the marked result, click "S" button on keyboard to ask the tool to save the label file. If not, click "Q" button will let the tool quit the labelling function.
\begin{figure*}[h]
	\centering
	\begin{subfigure}[b]{0.23\linewidth}
		\includegraphics[width=\linewidth]{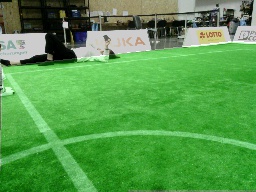}
		\caption{Raw Image}
		\label{fig:tool1}
	\end{subfigure}
	\begin{subfigure}[b]{0.23\linewidth}
		\includegraphics[width=\linewidth]{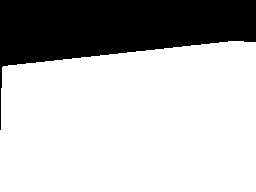}
		\caption{Mask}
		\label{fig:tool2}
	\end{subfigure}
	\begin{subfigure}[b]{0.23\linewidth}
		\includegraphics[width=\linewidth]{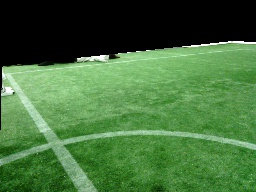}
		\caption{ROI}
		\label{fig:tool3}
	\end{subfigure}
	\begin{subfigure}[b]{0.23\linewidth}
		\includegraphics[width=\linewidth]{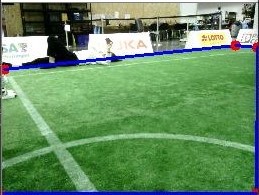}
		\caption{Points Marked}
		\label{fig:tool4}
	\end{subfigure}
	\caption{A sample labelling process}
	\label{fig:tool}
\end{figure*}


\section{Usage Example}
This dataset can be utilised in different vision applications. Mainly, this dataset is designed to help solving the field segmentation problem under various conditions. Machine learning algorithms as well as handcrafted vision algorithms can both utilise these data for development or validation. \\Videos are recorded under a variety of light conditions. This feature allows user to work on a colour recognition algorithm to find particular colour, such as \textbf{Green} for the field or \textbf{White} for robots, balls and lines.\\
Field features are also included in some images, such as corners, center circles and lines. Using the provided tool, user can generate their own Field Feature dataset. Since 2019, a statistical learning named Class Conditional Gaussian Mixture Model\cite{ashar} has been used by rUNSWift for detecting the field features where parts of the data used are generated by this dataset and tool.

\section{Conclusion}
In this paper, we propose a dataset collected under different environment condition. This dataset can be used for developing SPL robot vision algorithms especially soccer field segmentation.

\newpage
\begin{appendices}
\section{Individual Statistics}
\label{appendix:stats2}

\begin{figure}[H]\ContinuedFloat
	\centering
	
	\begin{subfigure}[b]{\textwidth}
		\includegraphics[width=\textwidth, height=10cm]{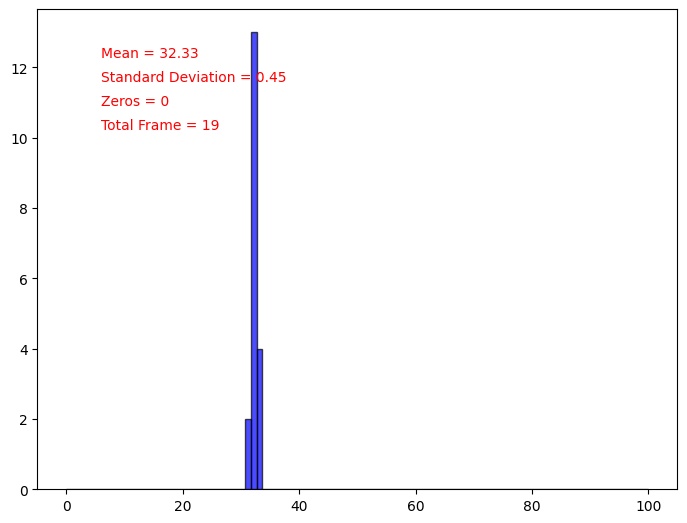}
		\caption{Statistics for Video index 1}
	\end{subfigure}
	\begin{subfigure}[b]{\textwidth}
		\includegraphics[width=\textwidth, height=10cm]{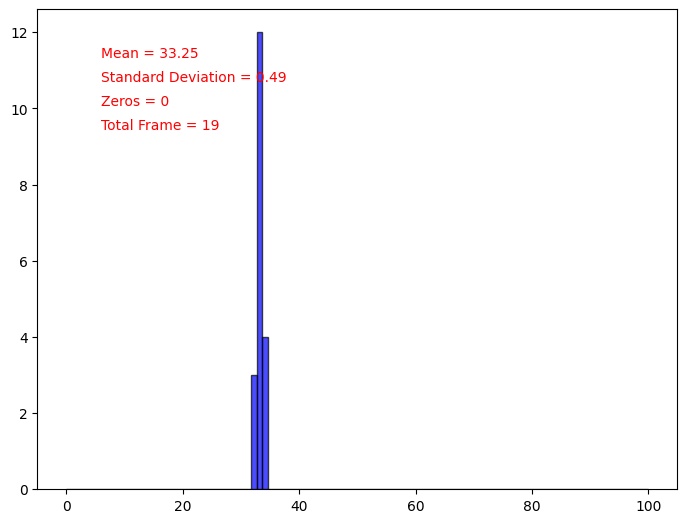}
		\caption{Statistics for Video index 2}
	\end{subfigure}
\end{figure}
\begin{figure}[H]\ContinuedFloat
		\begin{subfigure}[b]{\textwidth}
		\includegraphics[width=\textwidth, height=10cm]{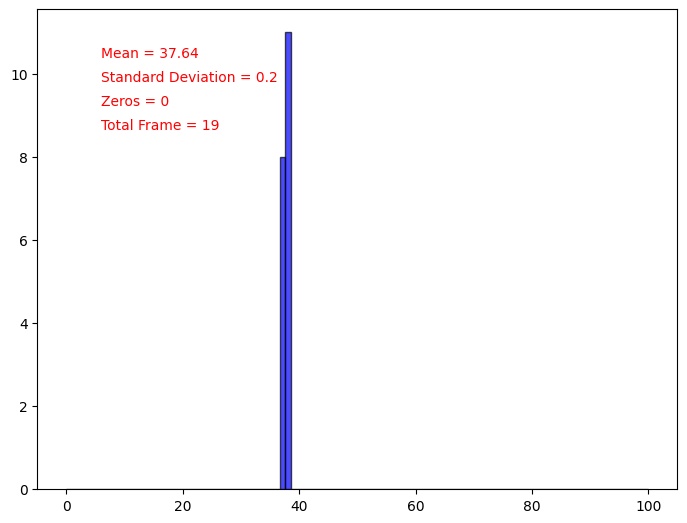}
		\caption{Statistics for Video index 3}
	\end{subfigure}
		\begin{subfigure}[b]{\textwidth}
		\includegraphics[width=\textwidth, height=10cm]{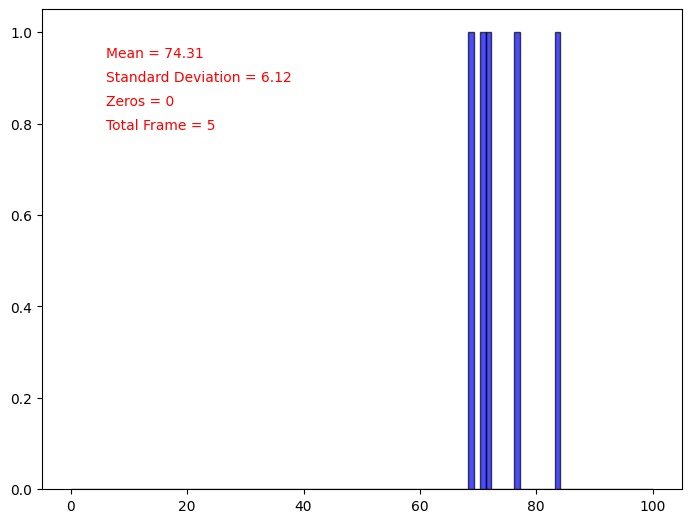}
		\caption{Statistics for Video index 4}
	\end{subfigure}
\end{figure}
\begin{figure}[H]\ContinuedFloat
	\begin{subfigure}[b]{\textwidth}
		\includegraphics[width=\textwidth, height=10cm]{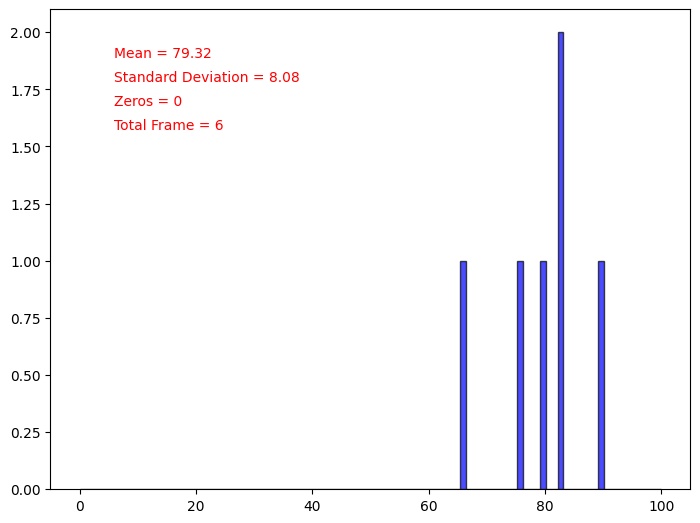}
		\caption{Statistics for Video index 5}
	\end{subfigure}
	\begin{subfigure}[b]{\textwidth}
		\includegraphics[width=\textwidth, height=10cm]{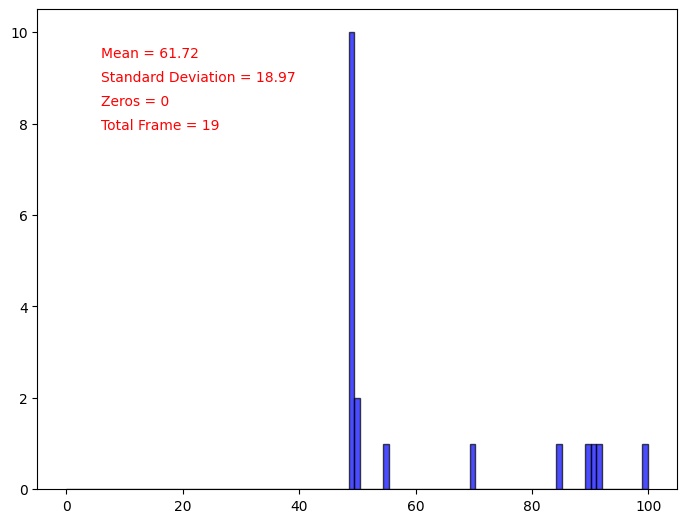}
		\caption{Statistics for Video index 7}
	\end{subfigure}
\end{figure}
\begin{figure}[H]\ContinuedFloat
	\begin{subfigure}[b]{\textwidth}
		\includegraphics[width=\textwidth, height=10cm]{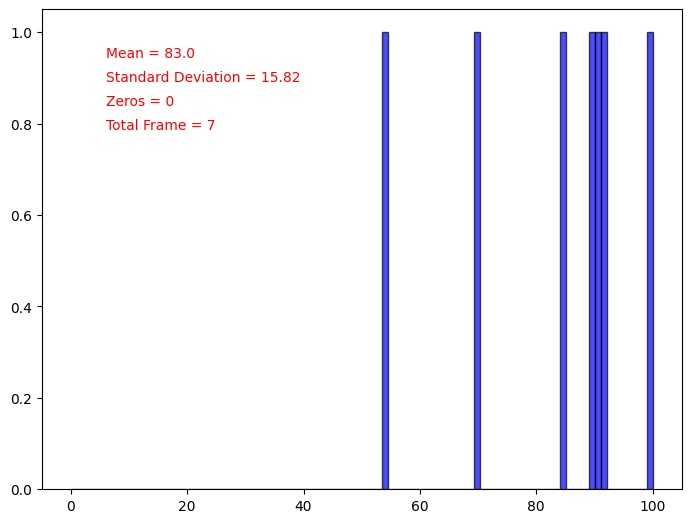}
		\caption{Statistics for Video index 8}
	\end{subfigure}
	\begin{subfigure}[b]{\textwidth}
		\includegraphics[width=\textwidth,  height=10cm]{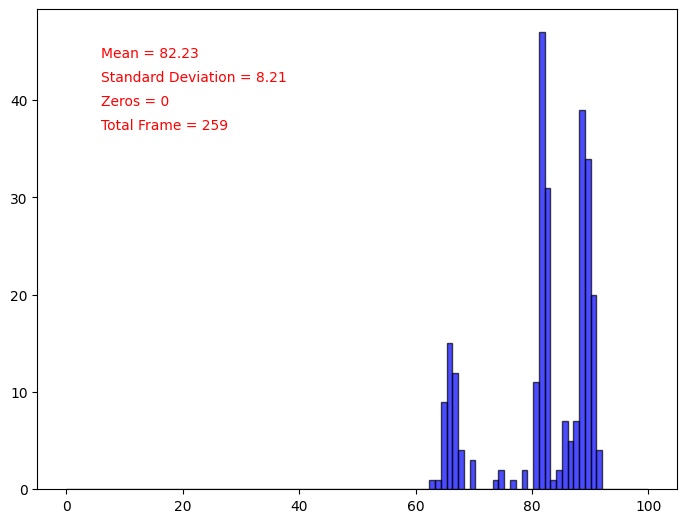}
		\caption{Statistics for Video index 9}
	\end{subfigure}
\end{figure}
\begin{figure}[H]\ContinuedFloat
	
	\begin{subfigure}[b]{\textwidth}
		\includegraphics[width=\textwidth height=10cm]{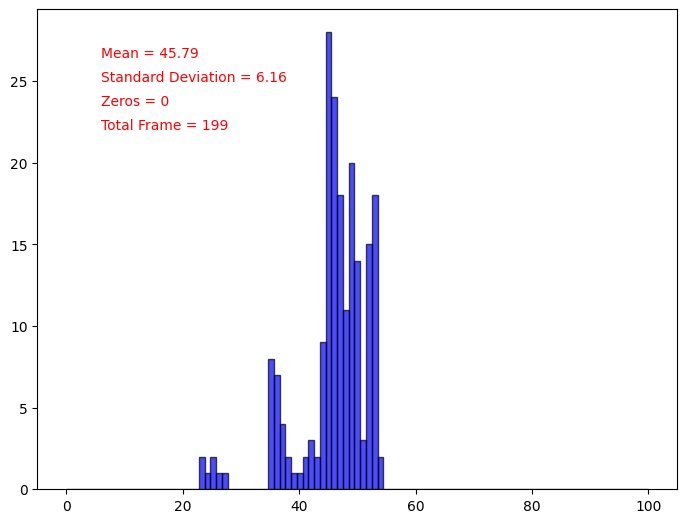}
		\caption{Statistics for Video index 10}
	\end{subfigure}
	\begin{subfigure}[b]{\textwidth}
		\includegraphics[width=\textwidth height=10cm]{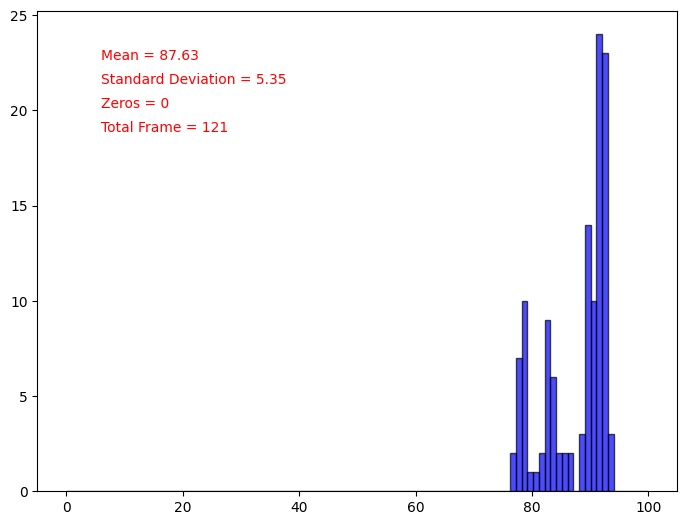}
		\caption{Statistics for Video index 11}
	\end{subfigure}
	\caption{}
\end{figure}
\begin{figure}[H]\ContinuedFloat
	
	\begin{subfigure}[b]{\textwidth}
		\includegraphics[width=\textwidth height=10cm]{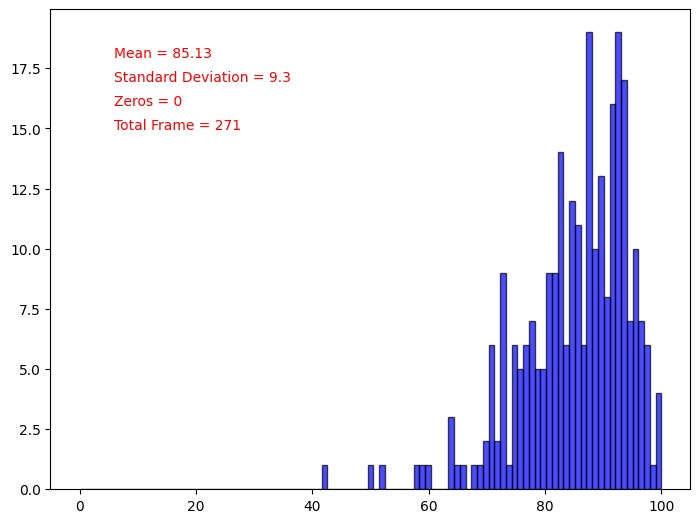}
		\caption{Statistics for Video index 12}
	\end{subfigure}
	\begin{subfigure}[b]{\textwidth}
		\includegraphics[width=\textwidth height=10cm]{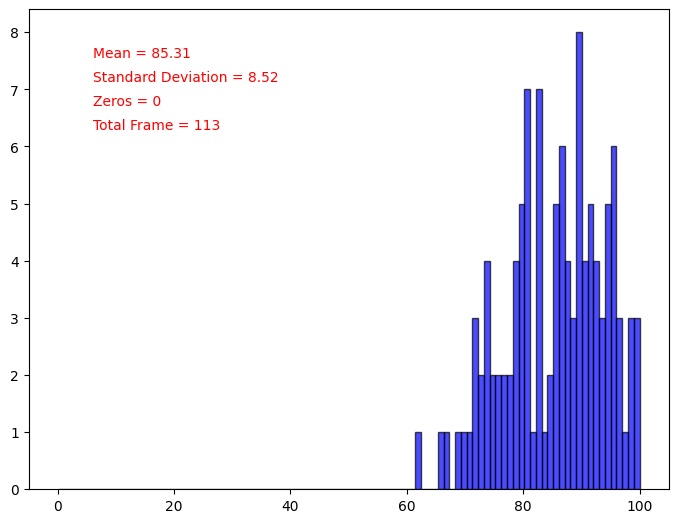}
		\caption{Statistics for Video index 13}
	\end{subfigure}
	\caption{}
\end{figure}
\begin{figure}[H]\ContinuedFloat
	
	\begin{subfigure}[b]{\textwidth}
		\includegraphics[width=\textwidth height=10cm]{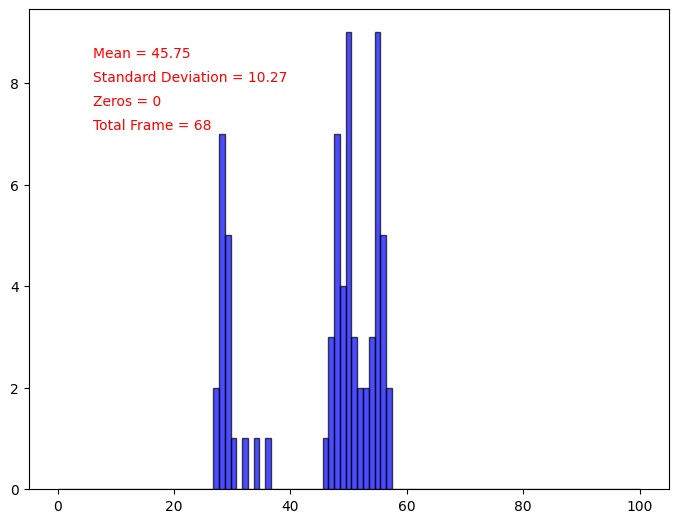}
		\caption{Statistics for Video index 14}
	\end{subfigure}
	\begin{subfigure}[b]{\textwidth}
		\includegraphics[width=\textwidth height=10cm]{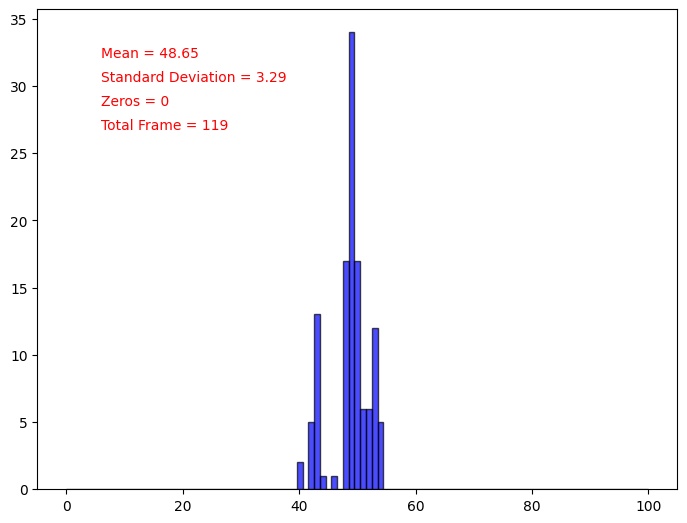}
		\caption{Statistics for Video index 15}
	\end{subfigure}
	\caption{}
\end{figure}
\begin{figure}[H]\ContinuedFloat
	
	\begin{subfigure}[b]{\textwidth}
		\includegraphics[width=\textwidth height=10cm]{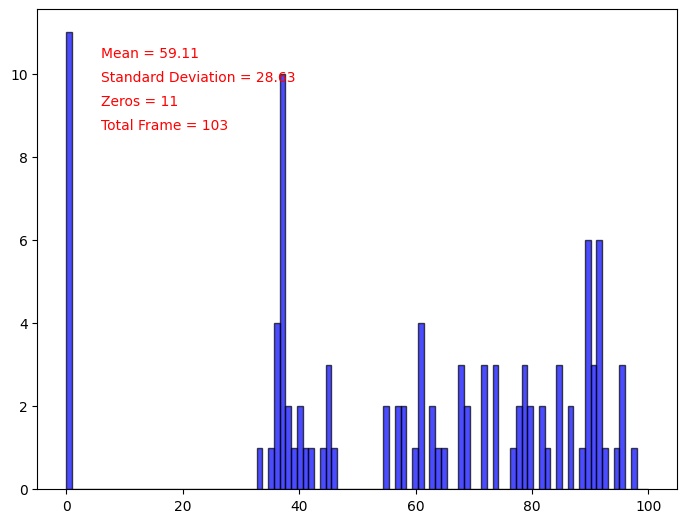}
		\caption{Statistics for Video index 16}
	\end{subfigure}
	\begin{subfigure}[b]{\textwidth}
		\includegraphics[width=\textwidth height=10cm]{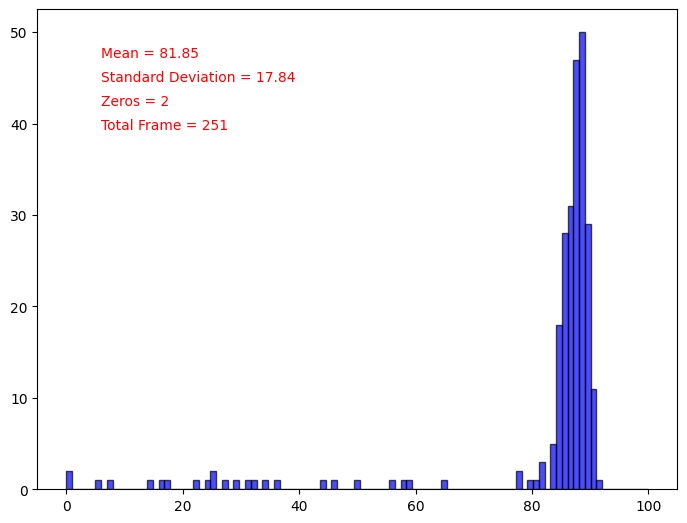}
		\caption{Statistics for Video index 17}
	\end{subfigure}
	\caption{}
\end{figure}
\begin{figure}[H]\ContinuedFloat
	
	\begin{subfigure}[b]{\textwidth}
		\includegraphics[width=\textwidth height=10cm]{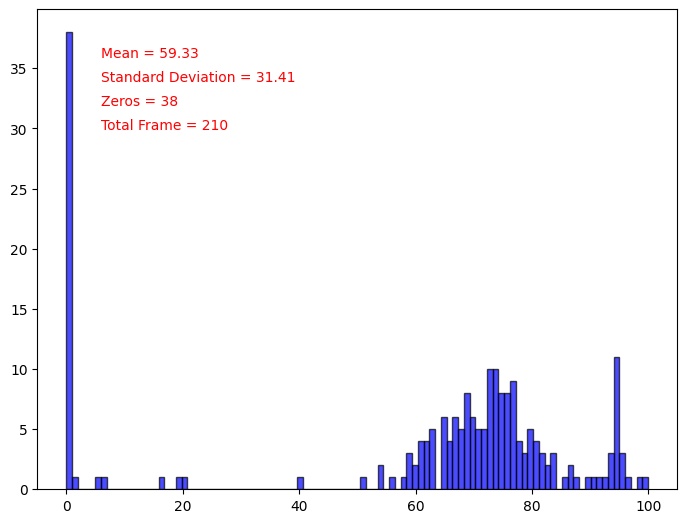}
		\caption{Statistics for Video index 18}
	\end{subfigure}
	\begin{subfigure}[b]{\textwidth}
		\includegraphics[width=\textwidth height=10cm]{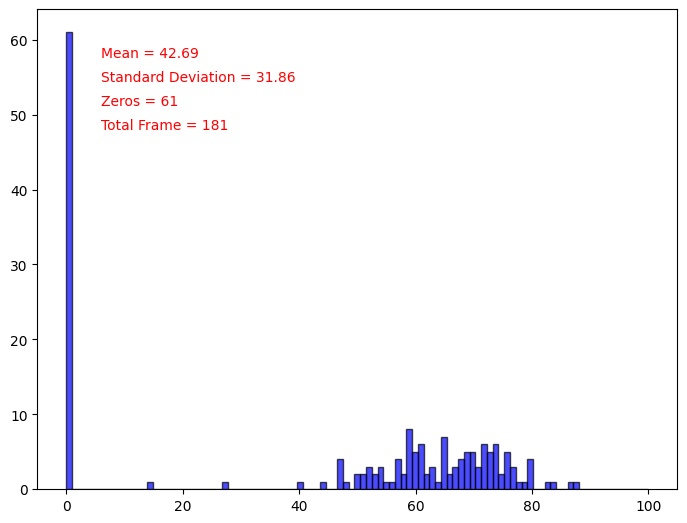}
		\caption{Statistics for Video index 19}
	\end{subfigure}
	\caption{}
\end{figure}
\begin{figure}[H]\ContinuedFloat
	
	\begin{subfigure}[b]{\textwidth}
		\includegraphics[width=\textwidth height=10cm]{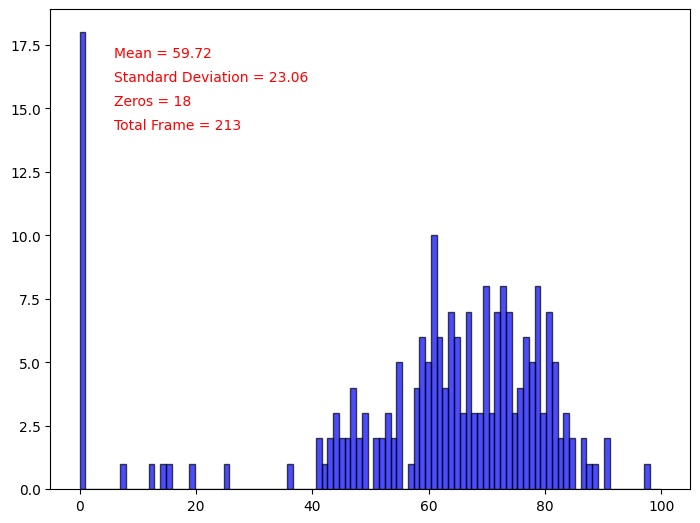}
		\caption{Statistics for Video index 20}
	\end{subfigure}
	\begin{subfigure}[b]{\textwidth}
		\includegraphics[width=\textwidth height=10cm]{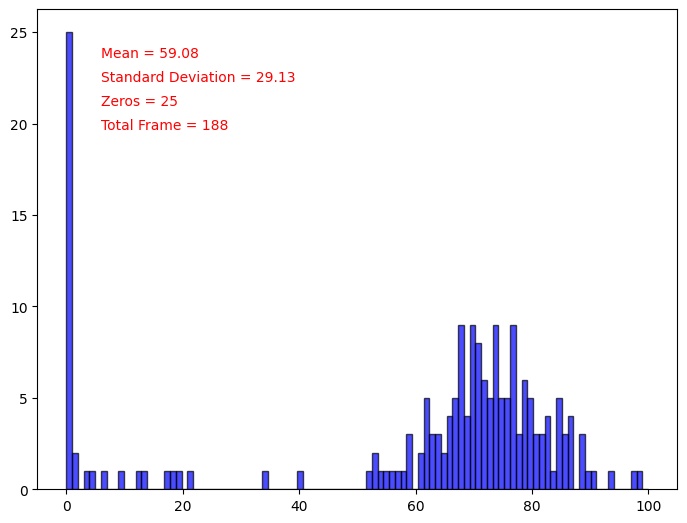}
		\caption{Statistics for Video index 21}
	\end{subfigure}
	\caption{}
\end{figure}

\end{appendices}
\bibliographystyle{acm}
\bibliography{main}

\end{document}